\documentclass[11pt]{article}

\usepackage[final]{acl}

\usepackage{times}
\usepackage{latexsym}

\usepackage[T1]{fontenc}
\usepackage[utf8]{inputenc}
\usepackage[T1]{fontenc}
\usepackage{amsmath, amssymb, amsfonts}
\usepackage{microtype}
\usepackage{hyperref}
\usepackage{xcolor}
\usepackage{graphicx}
\usepackage{natbib}
\usepackage{booktabs}
\usepackage{float}
\usepackage[table]{xcolor}
\usepackage[most]{tcolorbox}
\usepackage{xcolor}

\newcommand{\dyjsonkey}[1]{\textcolor{blue!60!black}{\textbf{"#1"}}}
\newcommand{\dyjsonval}[1]{\textcolor{black}{"#1"}}

\newtcolorbox{dypromptbox}[1]{
    colback=gray!5,
    colframe=gray!60,
    arc=2mm,
    boxrule=0.5pt,
    title=\textbf{#1},
    breakable,
    enhanced,
    fontupper=\small\ttfamily,
    before upper={\raggedright}
}
\definecolor{ivlband}{HTML}{E9F5EC}  
\definecolor{lovband}{HTML}{F3ECF7}  

\definecolor{propband}{HTML}{F2F2F2}
\definecolor{q25band}{HTML}{FFF5E6}
\definecolor{q3band}{HTML}{E6F0FA}
\definecolor{oursband}{HTML}{FFECEC}

\newcommand{\GRPOlat}{\mathrm{GRPO}_{\mathrm{latent}}}
\newcommand{\methodname}{DyLaR}  
\newcommand{\second}[1]{\underline{#1}}

\usepackage[utf8]{inputenc}

\usepackage{microtype}

\usepackage{inconsolata}

\usepackage{graphicx}

\title{Perception Before Reasoning: Dynamic Latent Reasoning for Video Understanding and Question Answering}

\author{Haotian Xia\textsuperscript{1}  
Zilin Xiao\textsuperscript{1}   
Junbo Zou\textsuperscript{2} 
Vicente Ordonez\textsuperscript{1} 
Hanjie Chen\textsuperscript{1}\\
\textsuperscript{1}Department of Computer Science, Rice University\\
\textsuperscript{2}College of Sciences, Georgia Institute of Technology\\
  \texttt{\small\{hx50, hanjie\}@rice.edu}
}

\begin{document}
\maketitle

\begin{abstract}
Video question answering requires models to ground language queries in visual evidence and, when necessary, reason over that evidence across time. 
Existing methods typically rely on long textual chain-of-thought rationales, even though many questions can be answered as soon as the relevant object, action, or frame is localized.
We propose Dynamic Latent Reasoning (\textbf{\methodname{}}), which first grounds a question in a short block of \emph{perception latents}—continuous hidden states that encode query-relevant visual evidence—and then adaptively decides whether to append \emph{reasoning latents}—continuous thoughts that reason over this evidence in latent space—before answering. 
\methodname{} learns this behavior by grounding perception latents in verified visual evidence and distilling verified rationales into reasoning latents, followed by reinforcement learning that further refines when to reason.
Across nine video benchmarks and four multimodal language model backbones, \methodname{} improves average accuracy over same-backbone baselines while generating fewer than 20 tokens per query. On Qwen3-VL-4B, for example, \methodname{} improves average accuracy over Qwen3-VL-4B-Thinking from 54.0 to 58.2 while reducing response length from 1,220.7 to 18.5 tokens per query.
Ablations further show that grounded perception latents, rationale-supervised reasoning latents, and adaptive routing each improve accuracy.

\end{abstract}

\section{Introduction}
\label{sec:intro}

Multimodal large language models (MLLMs) have shown strong capabilities in video understanding and reasoning~\citep{maaz2024video,lin2024video,zhang2025videollama}.
Recent advances further improve video reasoning with chain-of-thought supervision, reinforcement learning (RL) with verifiable rewards, and explicit evidence grounding~\citep{NEURIPS2025_8eb39768,wang2025videorftincentivizingvideoreasoning,meng2025open}. 
In particular, prevailing methods elicit a textual chain-of-thought (CoT) before the answer for every question~\citep{wei2022chain,NEURIPS2025_8eb39768}: the model first describes the relevant video content and then reasons over the description step by step, which improves accuracy through additional test-time computation~\citep{snell2024scalingllmtesttimecompute}, but at the cost of generating hundreds or thousands of tokens per query.
\begin{figure*}[h]
\centering
\includegraphics[width=0.8\textwidth]{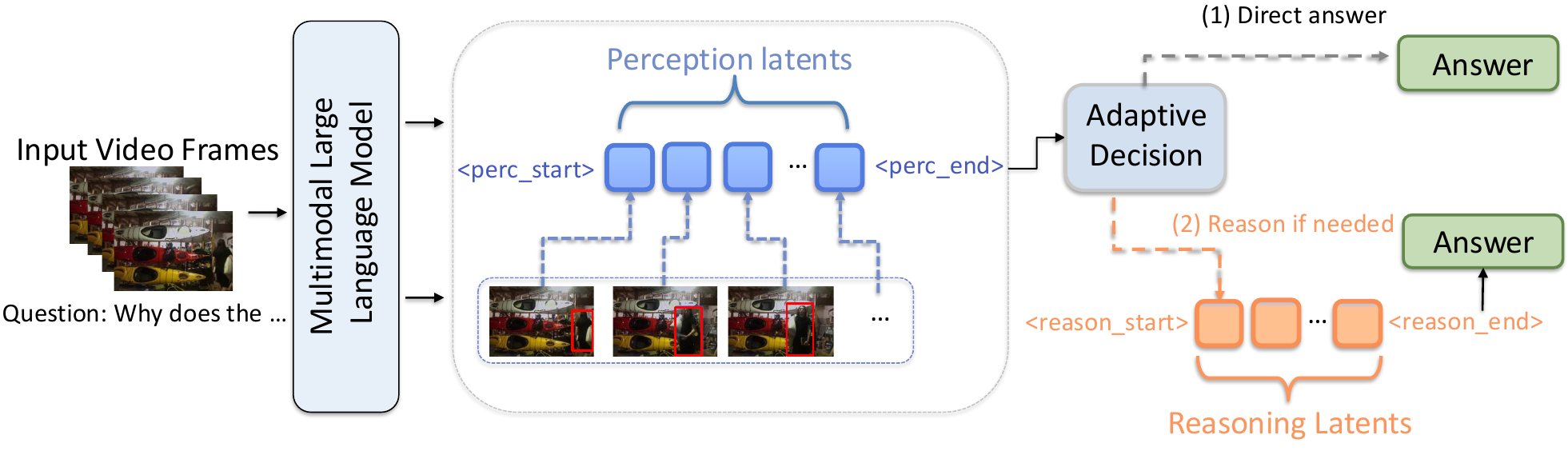}
\caption{
Overview of \methodname{}, an evidence-first latent response format for video question answering. 
The model first grounds the question with perception latents that encode query-relevant visual evidence. 
It then adaptively chooses between a direct answer path and a reasoning path that inserts additional reasoning latents before the final answer. 
By allocating reasoning latents only when needed, \methodname{} avoids unnecessary textual chain-of-thought generation while preserving answer-relevant visual evidence.
}
\label{fig:overview}
\end{figure*}

Latent reasoning offers a way to keep this benefit without paying explicit CoT's token cost.
A textual CoT's benefit comes from the hidden states it produces, not the words they are decoded into, so this computation can be carried out directly in continuous hidden states~\citep{hao2025traininglargelanguagemodels}.
Explicit CoT can even be distilled into a few continuous hidden states without losing accuracy~\citep{shen2025codi}.
In the image domain, recent work extends this idea to visual content: latent tokens are used to reconstruct query-relevant visual representations~\citep{lee2025lvr,zhang2025latentsketchpadsketchingvisual}, or latent visual states are interleaved with an explicit CoT~\citep{dong2026interleavedlatentvisualreasoning}, which still generates the full textual rationale in between.
These results suggest that visual evidence and the reasoning over it can be represented in latent space, preserving accuracy at a lower generation cost.

However, existing latent reasoning methods are designed for images and focus on how to reason in latent space, without considering whether reasoning is needed for a given question.
In video QA, we argue that whether reasoning is needed differs from question to question. 
For \emph{perception-oriented} questions, the answer is determined as soon as the relevant object, action, or frame is localized across the video; a textual rationale mostly re-describes what has already been found and is thus unnecessary. 
For \emph{reasoning-oriented} questions, localized evidence alone is insufficient: the model must further infer over it, for example comparing events across time or combining cues scattered in different frames. 
Every video question thus needs evidence grounding over multiple frames, but only some need reasoning on top of it. 
This motivates an adaptive latent response. 
A latent grounding stage captures query-relevant visual evidence for every question, and a latent reasoning stage infers over this evidence only when needed.

We propose Dynamic Latent Reasoning (\textbf{\methodname{}}), which first grounds a question in latent visual evidence and then adaptively decides whether additional latent reasoning is needed (Figure~\ref{fig:overview}).
Given a video and a question, \methodname{} first generates a short segment of \emph{perception latents}, which are continuous hidden states that carry query-relevant visual evidence. 
The next generated token then serves as an adaptive decision point: the model either answers directly from the perception latents, or extends the sequence with \emph{reasoning latents}, which are continuous thoughts that reason over the grounded evidence in latent space, before answering.
Rather than assigning the same reasoning budget to every question, \methodname{} keeps the grounded evidence in perception latents and allocates reasoning latents only when further inference over it is needed.
It learns this response format by grounding perception latents in verified, localized visual evidence and distilling verified rationales into reasoning latents, followed by reinforcement learning with verifiable rewards that further refines the decision of when to reason.

Experiments show that \methodname{} improves both accuracy and efficiency for video QA.
On the Qwen3-VL-4B backbone~\citep{bai2025qwen3vltechnicalreport}, \methodname{} improves average accuracy across nine benchmarks over Qwen3-VL-4B-Thinking from 54.0 to 58.2 while generating only 18.5 tokens per query, compared with 1,220.7 tokens from Qwen3-VL-4B-Thinking.
On the Qwen2.5-VL-7B backbone~\citep{bai2025qwen25vltechnicalreport}, \methodname{} achieves the best average accuracy among recent video reasoning models built on the same backbone~\citep{NEURIPS2025_8eb39768,wang2025videorftincentivizingvideoreasoning,meng2025open,liu2026videoautor1videoautoreasoning}, while using only 18.2 tokens per query.
The proposed method further improves InternVL3.5-4B~\citep{wang2025internvl35advancingopensourcemultimodal} and LLaVA-OneVision-7B~\citep{li2024llavaonevisioneasyvisualtask}, demonstrating that \methodname{} generalizes well across different VLM backbones.
Ablations show that removing perception grounding, reasoning latents, rationale-to-latent supervision, or adaptive routing all reduce accuracy, confirming that learning \emph{when} to reason matters as much as reasoning itself; attention-map analysis further shows that \methodname{} concentrates attention on query-relevant visual cues more than CoT baselines do, offering a non-textual explanation of the evidence behind each answer.

Our contributions are threefold:
\begin{itemize}
    \item We formulate video QA as adaptive latent reasoning: every question receives latent evidence grounding, while reasoning is invoked only when needed. 
    The model first grounds the question with perception latents, and then chooses between direct answering and latent reasoning before producing the answer.

    \item We introduce \methodname{}, a latent response framework that separates perception latents for grounded visual evidence from optional reasoning latents that infer over the grounded evidence, and train the two latent phases with localized evidence supervision, rationale-to-latent distillation, and verifiable reinforcement learning.

    \item Across nine video benchmarks and four backbones, \methodname{} achieves the best backbone-controlled average accuracy with fewer than 20 generated tokens per query. Ablations verify that perception grounding, reasoning-latent supervision, and adaptive routing each contribute, and attention analysis shows that the latent response concentrates on query-relevant visual evidence, providing a non-textual form of interpretability.
\end{itemize}

\section{Related Work}
\label{sec:related}

\paragraph{Video understanding and reasoning with MLLMs.}
Video MLLMs have progressed from video dialogue and instruction tuning, such as Video-ChatGPT, Video-LLaVA, LLaVA-Video, and VideoLLaMA3, to video reasoning models~\citep{maaz2024video,lin2024video,zhang2025llavavideovideoinstructiontuning,zhang2025videollama}. 
Recent reasoning-oriented methods use CoT supervision and reinforcement learning to improve video understanding~\citep{NEURIPS2025_8eb39768,wang2025videorftincentivizingvideoreasoning,liu2026videoautor1videoautoreasoning}, while Open-o3-Video grounds reasoning with explicit timestamps and bounding boxes~\citep{meng2025open}. 
Other work explores frame-aware or agentic reasoning~\citep{wang2024videoagent, zhi2025videoagent2enhancingllmbasedagent, ge2025framemind, he2025framethinker, wang2025video, zou2026videobrainlearningadaptiveframe}.
However, these methods rely on long textual reasoning traces, explicit evidence-gathering procedures, or additional interaction steps.

\paragraph{Latent reasoning.}
Explicit CoT improves reasoning but introduces token overhead~\citep{wei2022chain}. 
Latent reasoning instead moves intermediate computation into continuous hidden states~\citep{chen2025reasoninglanguagecomprehensivesurvey}. 
Coconut feeds hidden states back as continuous thoughts, and CODI compresses explicit CoT into continuous states through self-distillation~\citep{hao2025traininglargelanguagemodels,shen2025codi}. 
SoftCoT, hybrid latent reasoning, and multimodal continuous thoughts further explore soft or continuous alternatives to discrete CoT generation~\citep{xu2025softcotsoftchainofthoughtefficient,yue2025hybridlatentreasoningreinforcement,pham2025multimodalchaincontinuousthought}. 

\paragraph{Latent visual reasoning.}
Recent work extends latent reasoning to visual representations, showing that continuous visual tokens or latent visual thoughts can improve fine-grained visual reasoning~\citep{bigverdi2024perceptiontokensenhancevisual,lee2025lvr,wu2026lavitaligninglatentvisual,qin2025chainofvisualthoughtteachingvlmsthink,yang2025machinementalimageryempower,liu2026reasoningminddynamicmultimodal}. 
LVR reconstructs query-relevant visual tokens in latent space~\citep{lee2025lvr}, ILVR interleaves text generation with latent visual states supervised by features from helper images~\citep{dong2026interleavedlatentvisualreasoning}, and Latent Sketchpad generates visual latents that can be decoded into interpretable sketches~\citep{zhang2025latentsketchpadsketchingvisual}. 
Mull-Tokens inserts a fixed block of modality-agnostic latent tokens before the answer~\citep{ray2026mulltokensmodalityagnosticlatentthinking}; while trained mainly on image, it also transfers to video QA.

However, these methods either operate in image-centric or helper-image settings, or allocate the same latent budget to every question. 
Video QA instead requires the model to first localize question-relevant evidence across frames, while only some questions need evidence further composed through temporal, relational, or causal reasoning. 
This calls for a response space that separates visual grounding from the optional need for additional reasoning.

\section{Method}
\label{sec:method}


We introduce \textbf{\methodname{}}, an evidence-first framework that unifies visual grounding and conditional reasoning within a \emph{latent response}, where we keep ordinary text tokens for segment delimiters (special tokens) and the final answer, and continuous latent positions carry the intermediate hidden-state computation. 
The full sequence is decoded autoregressively by the same model.


Given a video and a question, the model first grounds the query in the visual input using a small number of \emph{perception latents}, which are continuous hidden states that encode query-relevant visual evidence. 
Conditioned on these latents, it then either produces the answer directly or, when the question demands further inference, inserts \emph{reasoning latents}, which are continuous thoughts that reason over the grounded evidence in latent space, before answering. 
In this way, every question receives visual grounding, while reasoning is invoked only when needed.

\subsection{Adaptive Latent Response Generation}
\label{sec:latent-generation}

Given a video-question pair $(V,q)$, the base MLLM encodes the sampled frames and the question into a multimodal prefix. 
Conditioned on this prefix, \methodname{} generates a structured latent response: delimiter and answer tokens are generated as ordinary text tokens, while perception and reasoning steps are represented by continuous latent positions.

\paragraph{Latent-segment decoding.}
Each latent segment occupies a contiguous span of positions marked by a fixed number of \texttt{<|latent\_pad|>} placeholders, which reserve positions but carry no semantic content of their own. 
Inside a segment, instead of sampling a vocabulary token and feeding its token embedding as the next input, the model feeds the previous step's decoder hidden state directly as the next input embedding, so that computation proceeds entirely in the continuous hidden-state space. 
Once the predetermined number of latent steps has elapsed, decoding reverts to ordinary token-by-token generation; we refer to this procedure as \emph{latent-segment decoding}.

\paragraph{Response structure.}
We define a perception segment $\mathcal{P}_{K_p}$, a reasoning segment $\mathcal{R}_{K_r}$, an answer field $\mathcal{A}(a)$ as
\[
\begin{aligned}
\mathcal{P}_{K_p}
&=
[
\texttt{<perc\_start>}
~\mathbf{z}^p_{1:K_p}
\texttt{<perc\_end>}
]\\
\mathcal{R}_{K_r}
&=
[
\texttt{<reason\_start>}
~\mathbf{z}^{r}_{1:K_r}
\texttt{<reason\_end>}
]\\
\mathcal{A}(a)
&=
[
\texttt{<answer>}
~a~\texttt{</answer>}
].
\end{aligned}
\]

Here, $\mathbf{z}^p_{1:K_p}$ denotes $K_p$ perception latents, $\mathbf{z}^{r}_{1:K_r}$ denotes $K_r$ reasoning latents, and $a$ denotes the final answer. 
The placeholders inside both latent segments are written as \texttt{<|latent\_pad|>} in the serialized response; their roles are determined by the surrounding delimiters.

\paragraph{Adaptive routing.}
The response therefore takes one of two forms:
\begin{equation}
\begin{aligned}
\mathcal{S}^{\textsc{direct}}
&=
[
\mathcal{P}_{K_p},
\mathcal{A}(a)
], \\
\mathcal{S}^{\textsc{reason}}
&=
[
\mathcal{P}_{K_p},
\mathcal{R}_{K_r},
\mathcal{A}(a)
].
\end{aligned}
\label{eq:response_forms}
\end{equation}
The position after \texttt{<perc\_end>} serves as the adaptive decision point: conditioned on the question and the generated perception latents, the model either emits \texttt{<answer>} to answer from grounded visual evidence, or emits \texttt{<reason\_start>} and allocates $K_r$ reasoning latents to integrate that evidence before answering. 


\subsection{Learning Grounded Perception and Conditional Reasoning}
\label{sec:learning}

Each training example carries two types of verified annotations: localized evidence boxes for the queried visual evidence and, when the question requires inference, a verified rationale. 
Examples with only verified evidence are labeled \textsc{direct}, and those with both are labeled \textsc{reason}. These annotations are automatically generated and filtered by an independent verifier model; Section~\ref{sec:exp-setup} describes the data sources and scale, and Appendix~\ref{app:data} details the generation and verification process.

We optimize \methodname{} in two stages. 
Supervised learning shapes what each latent segment represents: perception latents are aligned with localized visual evidence, while reasoning latents are trained to capture reasoning information. 
Reinforcement learning then refines complete responses using rewards for response format and answer correctness.

\paragraph{Supervised grounding and rationale distillation.}
\label{sec:sft}

Let $s \in \{\textsc{direct}, \textsc{reason}\}$ denote the verified type of a training example, matching the two response forms in Eq.~\eqref{eq:response_forms}: a \textsc{direct} example carries verified localized evidence boxes, and a \textsc{reason} example additionally carries a verified rationale. 
The supervised fine-tuning (SFT) objective is
\begin{equation}
\begin{aligned}
\mathcal{L}_{\mathrm{SFT}}
&=\, 
\mathcal{L}_{\mathrm{text}}
+
\lambda_g \mathcal{L}_{\mathrm{ground}}\\
&+
\mathbb{I}[s=\textsc{reason}]
\left(
\lambda_e \mathcal{L}_{\mathrm{exp}}
+
\lambda_d \mathcal{L}_{\mathrm{distill}}
\right),
\end{aligned}
\label{eq:sft_loss}
\end{equation}
where $\mathcal{L}_{\mathrm{text}}$ is the standard language-modeling loss over non-latent text positions in the target response. 
It includes structural delimiters and final answer tokens, while \texttt{<|latent\_pad|>} positions are ignored.
The grounding term $\mathcal{L}_{\mathrm{ground}}$ encourages perception-latent hidden states to align with object-level visual targets constructed from verified evidence boxes. 
For \textsc{reason} examples, the explicit-rationale term $\mathcal{L}_{\mathrm{exp}}$ applies language-modeling supervision to the rationale-and-answer view, and the distillation term $\mathcal{L}_{\mathrm{distill}}$ matches the two views' hidden states at the answer position: after its reasoning latents, the model should arrive at the same internal state it would have reached by generating the full rationale.
$\lambda_g$, $\lambda_e$, and $\lambda_d$ are hyperparameters.

\paragraph{Perception grounding.}
For each verified evidence box, we construct an object-level visual target from the video patches it covers. 
Let $\mathcal{P}_i$ denote the set of patch indices inside the $i$-th evidence box, and let $\mathbf{x}^v_p \in \mathbb{R}^{d}$ be the projected visual embedding of patch $p$ in the language-model hidden space. 
The target for the $i$-th perception latent is the box-averaged embedding
\begin{equation}
\mathbf{u}_i
=
\frac{1}{|\mathcal{P}_i|}
\sum_{p \in \mathcal{P}_i}
\mathbf{x}^v_p,
\qquad
\mathbf{u}_i \in \mathbb{R}^{d}.
\label{eq:perc_target}
\end{equation}
We aggregate patches within each box because the evidence needed for video QA is usually an object, region, or an action, rather than an isolated patch.
Evidence boxes are ordered chronologically across frames.

During supervised learning, the number of perception latents follows the number of verified boxes in the example. 
Let $M_p$ be this number. 
We teacher-force the perception-latent positions with these visual targets: $\mathbf{u}_i$ is used as the input embedding at the corresponding perception-latent position, and the preceding hidden state is aligned to $\mathbf{u}_i$ through the grounding loss.

Let $\hat{\mathbf{z}}^p_i$ denote the decoder hidden state immediately preceding the $i$-th perception latent, the grounding loss is the average cosine distance
\begin{equation}
\mathcal{L}_{\mathrm{ground}}
=
\frac{1}{M_p}
\sum_{i=1}^{M_p}
\left[
1-\cos\left(\hat{\mathbf{z}}^p_i,\mathbf{u}_i\right)
\right].
\label{eq:ground_loss}
\end{equation}
At inference time, evidence boxes are not available; perception latents are generated using the latent-segment decoding rule described in Section~\ref{sec:latent-generation}.

\paragraph{Rationale-to-latent self-distillation.}
Reasoning latents do not have direct visual targets. 
Building on CODI's self-distillation framework~\citep{shen2025codi}, we adapt explicit-to-implicit CoT distillation to the reasoning segment of our latent response.  
For each \textsc{reason} example, the same model performs two forward passes. 
The \emph{teacher} pass spells out the verified rationale in text before the answer and is trained with a language-modeling loss over its rationale and answer tokens, 
$\mathcal{L}_{\mathrm{exp}} = -\sum_{t \in \mathcal{T}_{\mathrm{exp}}} \log \pi_\theta\!\left(w_t \mid V, q, w_{<t}\right)$, 
where $w_t$ is the $t$-th token of the teacher response, $\pi_\theta$ is the model's next-token distribution, and $\mathcal{T}_{\mathrm{exp}}$ indexes the rationale and answer positions. 
The \emph{student} pass follows the reasoning form in Eq.~\eqref{eq:response_forms}, replacing the rationale with $K_r$ reasoning latents.


To transfer the rationale information into the reasoning latents, we align the two passes at the answer position, i.e., the position that predicts the first answer token. 
Let $\mathbf{h}^{\mathrm{lat}}_{\ell}$ and $\mathbf{h}^{\mathrm{exp}}_{\ell}$ denote the layer-$\ell$ decoder hidden states at this position in the student and teacher passes, respectively. 
We define
\begin{equation}
\mathcal{L}_{\mathrm{distill}}
= 
\frac{1}{L}
\sum_{\ell=1}^{L}
\mathrm{SmoothL1}
\left(
\mathbf{h}^{\mathrm{lat}}_{\ell},
\mathrm{sg}
\left[
\mathbf{h}^{\mathrm{exp}}_{\ell}
\right]
\right),
\label{eq:distill_loss}
\end{equation}
where $L$ is the number of decoder layers and $\mathrm{sg}[\cdot]$ denotes stop-gradient.
$\mathrm{SmoothL1}$ is applied element-wise and averaged over hidden dimensions: for $\Delta=\mathbf{x}-\mathbf{y}$, 
$\mathrm{SmoothL1}(\mathbf{x},\mathbf{y})=\frac{1}{d}\sum_{j=1}^{d}\phi(\Delta_j)$, where $\phi(u)=\frac{1}{2}u^2$ if $|u|<1$ and $\phi(u)=|u|-\frac{1}{2}$ otherwise.
The student is never trained to imitate rationale tokens; the rationale supervises it only at the representation level.

\paragraph{Reinforcement Learning with Latent Replay}
\label{sec:rl}
SFT provides latent supervision under fixed target sequences, while inference depends on the model's own routing, whether to invoke the optional reasoning latents before answering.
We therefore refine complete rollouts with GRPO~\citep{shao2024deepseekmathpushinglimitsmathematical} using verifiable rewards for response-format validity and answer correctness.

Since latent tokens are continuous hidden states rather than sampled vocabulary tokens, they do not have token-level probabilities. 
We adopt latent-state replay following $\GRPOlat$~\citep{lee2025lvr}. 

For each policy-scored text token $y_{i,t}$ in rollout $o_i$, the importance ratio is
\begin{equation}
r_{i,t}(\theta)
=
\frac{
\pi_\theta
\left(
y_{i,t} \mid V,q,\tilde{\mathbf{H}}_i,y_{i,<t}
\right)
}{
\pi_{\theta_{\mathrm{old}}}
\left(
y_{i,t} \mid V,q,\tilde{\mathbf{H}}_i,y_{i,<t}
\right)
},
\label{eq:rl_ratio}
\end{equation}
where $\pi_\theta$ and $\pi_{\theta_{\mathrm{old}}}$ are the current and rollout-generating policies, and $\tilde{\mathbf{H}}_i$ denotes the latent hidden states recorded during the generation of rollout $o_i$ and replayed as fixed input embeddings when its text tokens are re-scored.
The token after \texttt{<perc\_end>} is also policy-scored: \texttt{<answer>} yields the direct form, while \texttt{<reason\_start>} yields the reasoning form.
For each prompt, we sample a group of $G$ rollouts $\{o_i\}_{i=1}^{G}$ from the old policy and optimize the standard clipped GRPO objective over policy-scored text tokens, with group-normalized advantages and a KL penalty toward the frozen SFT reference; the full objective is given in Appendix~\ref{app:grpo}.

\paragraph{Reward.}
For each rollout $o_i$, we use a format-gated reward:
\begin{equation}
R(o_i)
=
\mathbb{I}_{\mathrm{form}}(o_i)
\left(
\alpha_{\mathrm{form}}
+
\alpha_{\mathrm{acc}}
\mathbb{I}_{\mathrm{acc}}(o_i)
\right),
\label{eq:reward}
\end{equation}
where $\mathbb{I}_{\mathrm{form}}(o_i)$ checks whether the rollout follows valid response form in Eq.~\eqref{eq:response_forms}, and $\mathbb{I}_{\mathrm{acc}}(o_i)$ checks whether the extracted answer matches the ground truth. 
The coefficients $\alpha_{\mathrm{form}}$ and $\alpha_{\mathrm{acc}}$ control the weights of format validity and answer accuracy. 
Group-normalizing these rewards across the group yields the advantages $\hat{A}_i$ in the GRPO objective (Appendix~\ref{app:grpo}).
\section{Experiments}
\label{sec:exp}

\subsection{Experiment Details}
\label{sec:exp-setup}
\paragraph{Models.}
Our main model is trained from Qwen3-VL-4B-Instruct~\citep{bai2025qwen3vltechnicalreport}. 
To compare with existing Qwen2.5-VL-based video reasoning baselines, we also instantiate \methodname{} on Qwen2.5-VL-7B-Instruct~\citep{bai2025qwen25vltechnicalreport}. To further verify that our method generalizes across different model families, we additionally instantiate \methodname{} on InternVL3.5-4B~\citep{wang2025internvl35advancingopensourcemultimodal} and LLaVA-OneVision-7B~\citep{li2024llavaonevisioneasyvisualtask}. For all backbones, we freeze the visual encoder and multimodal projector, and update only the language model parameters.
\begin{table*}[t]
\centering
\caption{
Accuracy (\%) on nine video-QA benchmarks with 16-frame uniform sampling.
Tok/query is the average number of visible tokens generated or prefilled per question.
LongVB: LongVideoBench. LVR: LongVideo-Reason. TempC: TempCompass. VHolmes: Video-Holmes. VTT: Video-TT.
}
\label{tab:main}
\resizebox{0.85\textwidth}{!}{
\begin{tabular}{l ccccc cccc c r}
\toprule
\textbf{Model} & Video-MME & LVBench & MVBench & MMVU & LongVB & LVR & TempC & VHolmes & VTT & \textbf{Avg.} & \textbf{Tok/query} \\
\midrule
\rowcolor{propband}\multicolumn{12}{l}{\emph{Proprietary models}} \\
\rowcolor{propband}GPT-4o & 71.2 & 48.9 & -- & 67.4 & 66.7 & 60.7 & -- & -- & 46.6 & -- & -- \\
\rowcolor{propband}Gemini-1.5-Pro & 74.5 & 33.1 & -- & 65.4 & 64.0 & 69.3 & -- & -- & -- & -- & -- \\
\midrule
\rowcolor{q25band}\multicolumn{12}{l}{\emph{Qwen2.5-VL-7B backbone}} \\
\rowcolor{q25band}Qwen2.5-VL-7B-Instruct + CoT
& 45.0 & 23.6 & 56.9 & 57.4 & 36.0 & 66.7 & 69.2 & 26.5 & 32.7 & 46.0 & 206.1 \\
\rowcolor{q25band}Video-R1
& 57.4 & 35.2 & 63.8 & 63.8 & 55.0 & 69.9 & 70.2 & 41.0 & \textbf{42.4} & 55.4 & 415.2 \\
\rowcolor{q25band}VideoRFT
& 57.5 & 34.9 & 62.7 & \textbf{65.6} & 53.3 & 70.0 & 70.8 & \second{41.6} & 41.2 & 55.3 & 358.9 \\
\rowcolor{q25band}Open-o3-Video
& \second{59.3} & \second{37.5} & 63.7 & 65.1 & \second{57.1} & \second{72.2} & 70.7 & 40.3 & 39.6 & 56.2 & 162.1 \\
\rowcolor{q25band}VideoAuto-R1
& 58.9 & 36.0 & \textbf{68.8} & \second{65.4} & 55.9 & \second{72.2} & 71.0 & \textbf{46.9} & 37.7 & \second{57.0} & 40.1 \\
\rowcolor{q25band}Mull-Token
& 58.2 & 37.2 & 65.6 & 63.5 & 53.7 & \second{72.2} & \second{72.1} & 38.6 & \second{41.3} & 55.8 & \second{27.0} \\
\rowcolor{q25band}\textbf{\methodname{} (Qwen2.5-VL-7B)}
& \textbf{59.8} & \textbf{41.2} & \second{66.3} & 64.5 & \textbf{57.4} & \textbf{74.6} & \textbf{72.4} & 41.0 & 40.0 & \textbf{57.5} & \textbf{18.2} \\
\midrule
\rowcolor{q3band}\multicolumn{12}{l}{\emph{Qwen3-VL-4B backbone}} \\
\rowcolor{q3band}Qwen3-VL-4B-Instruct + CoT
& 47.1 & 26.0 & 52.7 & 62.7 & 41.7 & 62.6 & \second{70.3} & 30.3 & 31.3 & 47.2 & 3572.3 \\
\rowcolor{q3band}Qwen3-VL-4B-Thinking
& \second{56.3} & \second{36.9} & \second{61.0} & \textbf{66.7} & \second{56.3} & \second{66.8} & \textbf{71.3} & \second{36.7} & \second{34.0} & \second{54.0} & \second{1220.7} \\
\rowcolor{q3band}\textbf{\methodname{} (Qwen3-VL-4B)}
& \textbf{59.6} & \textbf{39.8} & \textbf{66.7} & \second{63.8} & \textbf{58.9} & \textbf{76.9} & 69.6 & \textbf{47.5} & \textbf{40.7} & \textbf{58.2} & \textbf{18.5} \\
\midrule
\rowcolor{ivlband}\multicolumn{12}{l}{\emph{InternVL3.5-4B backbone}} \\
\rowcolor{ivlband}InternVL3.5-4B-Instruct + CoT
& \second{54.5} & \second{37.4} & \second{63.1} & \second{62.2} & \second{51.8} & \second{69.5} & \second{69.5} & \second{38.6} & \textbf{39.8} & \second{54.0} & \second{173.9} \\
\rowcolor{ivlband}\textbf{\methodname{} (InternVL3.5-4B)}
& \textbf{57.6} & \textbf{38.5} & \textbf{67.2} & \textbf{62.7} & \textbf{56.8} & \textbf{73.9} & \textbf{70.7} & \textbf{44.7} & \second{38.2} & \textbf{56.7} & \textbf{13.4} \\
\midrule
\rowcolor{lovband}\multicolumn{12}{l}{\emph{LLaVA-OneVision-7B backbone}} \\
\rowcolor{lovband}LLaVA-OneVision-7B + CoT
& \second{53.0} & \second{34.0} & \second{56.9} & \second{53.0} & \second{50.6} & \second{66.2} & \second{65.4} & \second{30.2} & \second{37.7} & \second{49.7} & \textbf{14.1} \\
\rowcolor{lovband}\textbf{\methodname{} (LLaVA-OneVision-7B)}
& \textbf{55.7} & \textbf{37.3} & \textbf{61.0} & \textbf{56.6} & \textbf{55.8} & \textbf{75.1} & \textbf{66.0} & \textbf{42.0} & \textbf{38.0} & \textbf{54.2} & \second{17.8} \\
\bottomrule
\end{tabular}
}
\end{table*}

\paragraph{Training data.}

We curate training data from Video-R1-CoT~\citep{NEURIPS2025_8eb39768}, which aggregates LLaVA-Video-178k~\citep{zhang2025llavavideovideoinstructiontuning}, NeXT-QA~\citep{xiao2021next}, PerceptionTest~\citep{patraucean2023perception}, etc., and supplement it with the training splits of LongVideo-Reason~\citep{chen2025scalingrllongvideos}, CG-Bench~\citep{chen2024cgbenchcluegroundedquestionanswering}, Video-Holmes~\citep{cheng2025video} and MLVU~\citep{zhou2025mlvubenchmarkingmultitasklong}. 
For each example, we generate candidate evidence boxes and rationales with Qwen3.5-397B-A17B~\citep{qwen35blog} and filter them with Kimi-K2.5~\citep{kimiteam2026kimik25visualagentic} as an independent verifier (prompts in Appendix~\ref{app:data}; human audit in Appendix~\ref{app:human_verif}). 
We retain approximately 20K examples for SFT and 30K for RL.

\paragraph{Training and inference.}
We apply the same training and inference procedure across all four backbones: one epoch of SFT followed by 2{,}500 GRPO steps. 
For both training and inference, we uniformly sample 16 frames per video, with each frame capped at 307{,}200 pixels; at inference time we use greedy decoding with fixed latent budgets $K_p=4$ and $K_r=6$. 
Full hyperparameters are in Appendix~\ref{app:hyper}.

\paragraph{Benchmarks.}
We evaluate nine video-QA benchmarks covering both perception-centric and reasoning-centric questions: Video-MME~\citep{fu2025video}, LVBench~\citep{wang2025lvbench}, LongVideoBench~\citep{wu2024longvideobench}, MVBench~\citep{li2024mvbench}, LongVideo-Reason~\citep{chen2025scalingrllongvideos}, TempCompass~\citep{liu2024tempcompassvideollmsreally}, Video-TT~\citep{zhang2025videothinkingtestholistic}, Video-Holmes~\citep{cheng2025video}, and MMVU~\citep{zhao2025mmvumeasuringexpertlevelmultidiscipline}.
Details of benchmarks are provided in Appendix~\ref{app:benchmarks}.

\paragraph{Baselines.}

We compare with four groups of baselines. 
First, we include reported results from proprietary models, GPT-4o~\citep{openai2024gpt4ocard} and Gemini-1.5-Pro~\citep{geminiteam2024gemini15unlockingmultimodal}, as reference points for frontier model performance; these models may use different inference settings. 
Second, for the Qwen2.5-VL-7B backbone, we compare against Qwen2.5-VL-7B-Instruct with CoT prompting and recent video reasoning models built on the same backbone, including Video-R1~\citep{NEURIPS2025_8eb39768}, VideoRFT~\citep{wang2025videorftincentivizingvideoreasoning}, Open-o3-Video~\citep{meng2025open}, VideoAuto-R1~\citep{liu2026videoautor1videoautoreasoning}, and Mull-Token~\citep{ray2026mulltokensmodalityagnosticlatentthinking}. Third, for the Qwen3-VL-4B backbone, we compare against Qwen3-VL-4B-Instruct with explicit CoT prompting and Qwen3-VL-4B-Thinking~\citep{bai2025qwen3vltechnicalreport}. 
Fourth, for the InternVL3.5-4B and LLaVA-OneVision-7B backbones, we compare against their Instruct models with CoT prompting. Evaluation protocol details are in Appendix~\ref{app:benchmarks}.

\subsection{Main Results}
\label{sec:main-results}
\begin{table*}[t]
\centering
\small
\caption{
SFT objective ablation on LVBench with Qwen3-VL-4B, before RL.
``Perception latent + text CoT'' keeps the perception latent block but verbalises the rationale;
``Text-CoT SFT'' performs all reasoning in the vocabulary space;
``Answer-only SFT'' removes both latent blocks and supervises only the final answer.
}\label{tab:ablation_sft_lvb}
\resizebox{0.85\textwidth}{!}{
\begin{tabular}{l cccccc c}
\toprule
\textbf{Variant} & Entity & Event & Key Info. & Reason & Temporal & Summ. & \textbf{Overall} \\
\midrule
\multicolumn{8}{l}{\emph{Text-reasoning SFT baselines}} \\
Perception latent + text CoT
& 38.7 & \second{40.5} & 40.2 & 38.8 & 33.2 & 27.6 & 38.9 \\
Text-CoT SFT
& 40.2 & 37.6 & \second{40.9} & 35.3 & 32.7 & \second{32.8} & \second{39.0} \\
Answer-only SFT
& \second{40.6} & 36.6 & 39.9 & 39.8 & \textbf{36.8} & \second{32.8} & 38.7 \\
\midrule
\multicolumn{8}{l}{\emph{\methodname{} latent SFT}} \\
\textbf{\methodname{} SFT}
& \textbf{41.5} & \textbf{41.7} & 39.2 & 36.3 & \second{34.5} & \second{32.8} & \textbf{40.5} \\
\quad w/o reasoning latents
& 40.2 & 40.0 & 38.1 & 39.8 & \second{34.5} & 25.9 & 39.1 \\
\quad w/o reasoning supervision
& 38.0 & 37.7 & \textbf{42.3} & 35.3 & 32.7 & \second{32.8} & 38.3 \\
\quad w/o perception grounding
& 38.1 & 38.3 & 39.2 & \textbf{41.3} & 34.1 & \textbf{34.5} & 38.7 \\
\quad w/o adaptive routing (mandatory latent reasoning)
& 39.1 & 39.4 & 38.5 & \second{40.3} & 33.2 & 29.3 & 39.1 \\
\bottomrule
\end{tabular}
}
\end{table*}

Table~\ref{tab:main} reports results on nine video benchmarks. 
Proprietary models provide a scale reference, while the Qwen2.5-VL-7B, Qwen3-VL-4B, InternVL3.5-4B, and LLaVA-OneVision-7B groups compare models built on the same backbone.

On the Qwen3-VL-4B backbone, \methodname{} achieves the best average performance among the compared Qwen3 models, improving over Qwen3-VL-4B-Thinking by 4.2 points on average. 
Although Qwen3-VL-4B-Thinking remains stronger on MMVU and TempCompass, \methodname{} achieves the best overall average while generating only 18.5 visible tokens per query, compared with 1{,}220.7 tokens from Qwen3-VL-4B-Thinking and 3{,}572.3 tokens from explicit CoT prompting.

On the Qwen2.5-VL-7B backbone, \methodname{} achieves the best overall average accuracy while using significantly fewer visible tokens. 
It outperforms Video-R1 by 2.1 points and VideoRFT by 2.2 points on average, and also surpasses the strongest prior baseline, VideoAuto-R1. 
Meanwhile, \methodname{} uses only 18.2 tokens on average, compared with 27.0 for Mull-Token, 40.1 for VideoAuto-R1, 162.1 for Open-o3-Video, 415.2 for Video-R1, and 358.9 for VideoRFT. Qualitative example visualizations can be found in Appendix~\ref{app:vis}.

The same pattern generalizes to the other two backbone families. 
On InternVL3.5-4B, \methodname{} improves the average from 54.0 to 56.7 while reducing generated tokens from 173.9 to 13.4 tokens per query. 
On LLaVA-OneVision-7B, it improves the average from 49.7 to 54.2. Its CoT baseline emits few visible tokens (14.1) due to weak adherence to the CoT instruction, yet \methodname{} still gains 4.5 points at a comparable budget (17.8).
Together with the two Qwen families, this shows that the evidence-first latent response transfers across four backbones rather than exploiting a property of a single model family.

\begin{figure*}[h!]
\centering
\includegraphics[width=0.78\textwidth]{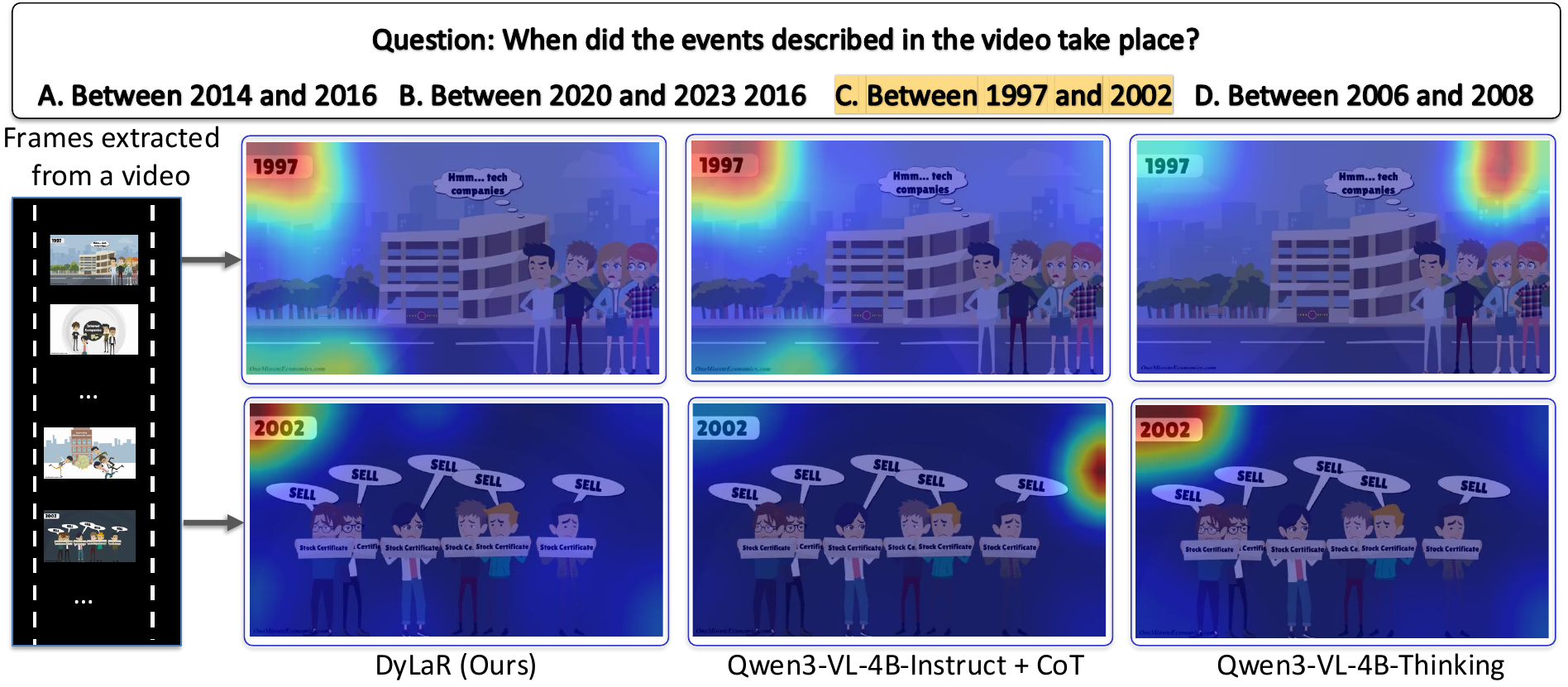}
\caption{
Case study. 
Heatmaps show Gaussian-smoothed attention aggregation over video patches for a temporal QA example. 
The example requires identifying the event period from two visual cues, ``1997'' and ``2002''. 
\methodname{} focuses on the diagnostic cues across frames, while CoT-based baselines show more diffuse or shifted attention. 
}
\label{fig:case_viz}
\end{figure*}

\subsection{Ablation Studies and Analysis}
\label{sec:ablation}

\paragraph{SFT objective ablation.}
Table~\ref{tab:ablation_sft_lvb} shows which supervision signals are needed to initialize the adaptive text-latent response. 
The three text-reasoning baselines (defined in Table~\ref{tab:ablation_sft_lvb}) all trail the full \methodname{} SFT, showing that neither explicit rationales nor an unsupervised latent architecture alone reproduces this behavior.

Within the \methodname{} family, each component matters: removing reasoning latents drops the score to 39.1 (visual grounding alone cannot compose evidence across frames); removing rationale supervision drops it further to 38.3 (reasoning latents need explicit rationale-to-latent supervision, not just extra computation); removing perception grounding drops it to 38.7 (perception latents need grounding in localized evidence); and replacing adaptive routing with mandatory latent reasoning -- always allocating $K_r$ latents regardless of difficulty -- also drops it to 39.1, showing that learning when to reason beats reasoning indiscriminately.
We also examine 32- and 64-frame budgets, where the full \methodname{} objective remains the best variant; details are in Appendix~\ref{app:sft_frames}.

\begin{table}[t]
\centering
\small
\caption{
Effect of the RL stage on Qwen2.5-VL-7B.
Results for the other three backbones are in Appendix~\ref{RL_overall}.
}
\label{tab:rl}
\resizebox{0.85\columnwidth}{!}{
\begin{tabular}{l c c}
\toprule
\textbf{Benchmark} & \textbf{SFT} & \textbf{$+$ RL (\methodname{})} \\
\midrule
Video-MME & 58.7 & \textbf{59.8}\,{\scriptsize\textcolor{green!55!black}{$+$1.1}} \\
LVBench & 40.6 & \textbf{41.2}\,{\scriptsize\textcolor{green!55!black}{$+$0.6}} \\
MVBench & 65.6 & \textbf{66.3}\,{\scriptsize\textcolor{green!55!black}{$+$0.7}} \\
MMVU & 61.4 & \textbf{64.5}\,{\scriptsize\textcolor{green!55!black}{$+$3.1}} \\
LongVideoBench & 55.8 & \textbf{57.4}\,{\scriptsize\textcolor{green!55!black}{$+$1.6}} \\
LongVideo-Reason & 73.2 & \textbf{74.6}\,{\scriptsize\textcolor{green!55!black}{$+$1.4}} \\
TempCompass & 71.8 & \textbf{72.4}\,{\scriptsize\textcolor{green!55!black}{$+$0.6}} \\
Video-Holmes & 39.8 & \textbf{41.0}\,{\scriptsize\textcolor{green!55!black}{$+$1.2}} \\
Video-TT & 37.4 & \textbf{40.0}\,{\scriptsize\textcolor{green!55!black}{$+$2.6}} \\
\midrule
\textbf{Avg.} & 56.0 & \textbf{57.5}\,{\scriptsize\textcolor{green!55!black}{$+$1.5}} \\
\bottomrule
\end{tabular}
}
\end{table}
\paragraph{RL refinement.}
Table~\ref{tab:rl} compares the full SFT with the final RL-refined model on the Qwen2.5-VL-7B backbone. 
RL improves the average accuracy from 56.0 to 57.5. 
The other three backbones follow the same pattern: RL also improves the nine-benchmark average for all the other three backbones; details are in Appendix~\ref{RL_overall}.

\begin{figure}[t]
    \centering
    \includegraphics[width=0.85\columnwidth]{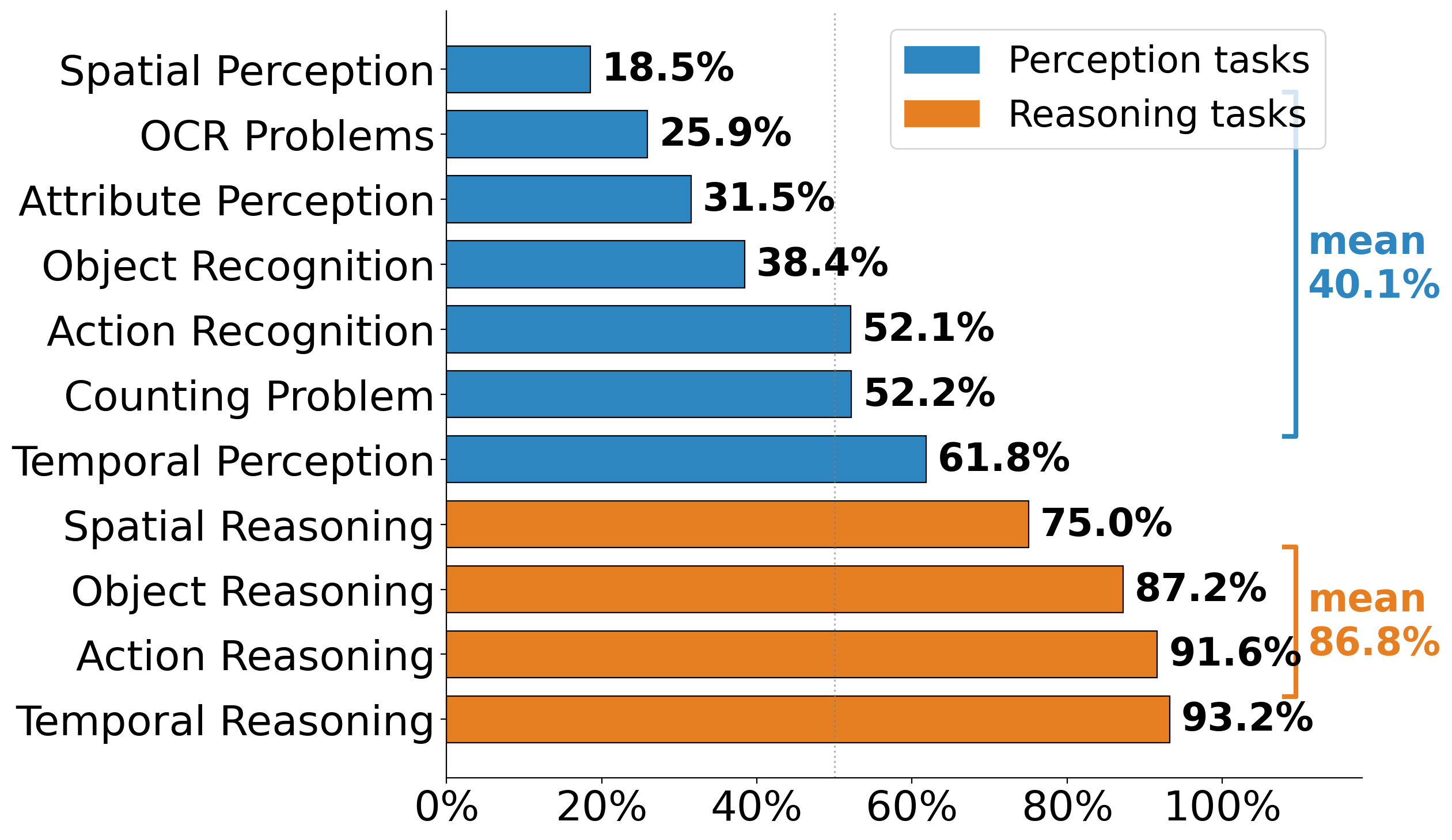}
    \caption{
    We report the fraction of examples on which \methodname{} chooses the optional reasoning branch.}
    \label{fig:routing_vdommb}
\end{figure}

\paragraph{Sensitivity to visual input budget.}
We test whether \methodname{}'s gains depend on a specific input configuration along two axes: the number of frames sampled at inference, and the pixel budget used for training and inference.
First, sampling 16, 32, or 64 frames at test time from the same \methodname{} (Qwen3-VL-4B) raises the nine-benchmark average from 58.2 to 59.7 to 60.5, and \methodname{} stays ahead of Qwen3-VL-4B-Thinking at every frame budget; details are in Appendix~\ref{app:frame_budget}.
Second, to show that the gains of \methodname{} are not simply due to the higher training pixel budget used in Table~\ref{tab:main}, we retrain \methodname{} (Qwen2.5-VL-7B) using the same lower pixel budget as Video-R1 and VideoRFT ($128\times28\times28$ for training and $256\times28\times28$ for evaluation). Under this setting, \methodname{} still achieves the highest average accuracy across nine benchmarks among the three methods (Appendix~\ref{app:lowres}), showing that the improvements consistently transfer across different pixel budgets.

\paragraph{Adaptive routing analysis.}
Figure~\ref{fig:routing_vdommb} tests whether \methodname{} reasons only when needed: under the official Video-MME category taxonomy, it triggers the reasoning branch on 40.1\% of examples across the seven perception categories but on 86.8\% across the four reasoning categories, showing that it reserves latent computation for reasoning-intensive questions rather than overusing the reasoning path.

\paragraph{Case analysis and visualization.}
Figure~\ref{fig:case_viz} visualizes a temporal QA example whose answer requires combining two temporal cues, ``1997'' and ``2002'', across frames. 
Compared with Qwen3-VL-4B-Instruct CoT and Qwen3-VL-4B-Thinking, \methodname{} produces heatmaps concentrated on these diagnostic cues and their supporting regions, suggesting that the latents preserve localized visual evidence before the answer decision. 
While the compact latent response forgoes a textual rationale, this attention concentration offers a complementary, non-textual form of interpretability.

\section{Conclusion}
\label{sec:conclusion}

We presented \methodname{}, an evidence-first latent response framework that grounds video questions with perception latents and conditionally allocates reasoning latents before answering. 
The model learns this response format from localized evidence supervision, rationale-to-latent distillation, and reinforcement learning with verifiable rewards. 
Across nine video benchmarks and four backbones, \methodname{} improves both accuracy and efficiency, and ablations confirm the value of both grounded perception states and latent reasoning.
\section*{Limitations}
Although \methodname{} reduces visible text generation by moving intermediate computation into latent states, these states are not directly human-readable, which limits the interpretability of individual reasoning steps. 
The attention-map analysis in Section~\ref{sec:ablation} partially mitigates this by exposing which visual evidence the model relies on before answering, though it does not recover a step-by-step textual rationale.
In addition, our experiments focus on four open-source MLLM backbones with uniform frame sampling; future work could study whether the same grounding--reasoning separation transfers to a broader set of architectures and adaptive frame-selection settings.

\section*{Ethical Considerations}
Our work builds on publicly released academic datasets and open-weight models, used consistently with their licenses and intended research use; our released code, annotations, and model checkpoints will follow the same terms, and we do not redistribute the original videos. The source benchmarks are curated academic datasets; our annotations add evidence boxes and rationales over them and introduce no personally identifying information or offensive content beyond what may exist in the original public datasets. The human audit of annotation quality was conducted by the authors. As with other video QA systems, our models can produce incorrect answers and should not be relied on in safety-critical settings.

\bibliography{custom}

\appendix
\section{Data Annotation and Verification Prompts}
\label{app:data}

This appendix provides the prompt templates used to construct and verify the training annotations for \methodname{}. 
The annotation pipeline is designed to provide two types of supervision: localized visual evidence for perception latents and, when additional inference is required, concise rationales for rationale-based latent supervision.

\subsection{Annotation Overview}

For each video-question pair, we uniformly sample 16 frames and use Qwen3.5-397B-A17B as the annotation generator. 
The generator produces a structured annotation containing a final answer, a set of question-relevant evidence boxes, and an optional rationale. 
We distinguish two answerable modes. 
In \textsc{direct} examples, the localized evidence directly supports the answer. 
In \textsc{reason} examples, the relevant evidence is visible but must be composed through temporal comparison, causal reasoning, relational reasoning, or option-level reasoning.

We then use Kimi-K2.5 as an independent verifier. 
The verifier checks whether the generated boxes contain useful evidence, whether the boxes support the gold answer, and, for reasoning examples, whether the generated rationale is faithful to the visible evidence. 
We discard malformed outputs, invalid boxes, unsupported samples, and annotations whose final answers disagree with the gold answer. 
Only verified examples are used for supervised fine-tuning.

\subsection{Annotation Generation Prompt}

The annotation generator is instructed to localize visual evidence before deciding whether the example should be treated as direct perception or reasoning. 
The simplified prompt template is shown below.

\begin{dypromptbox}{Annotation Generation Prompt}
You are an evidence-first video annotation assistant. Return strict JSON only.\\[0.4em]

Input:\\
- A video question.\\
- Candidate answer options, if available.\\
- 16 uniformly sampled video frames with frame indices and timestamps.\\[0.4em]

Task:\\
1. Annotate all question-relevant visual evidence in the visible frames.\\
\hspace*{1em}Evidence may include subjects, objects, actions, states, transitions, temporal bookends, or visual cues needed to answer the question.\\[0.2em]

2. Decide whether the current visual evidence is sufficient.\\[0.2em]

3. If the evidence directly supports the answer, use mode = \dyjsonval{perception\_only}.\\
\hspace*{1em}Provide evidence boxes, the final answer, and an empty rationale.\\[0.2em]

4. If the evidence is visible but requires comparison, temporal integration, causality, relational reasoning, or option elimination, use mode = \dyjsonval{need\_reasoning}.\\
\hspace*{1em}Provide evidence boxes, the final answer, and one concise rationale.\\[0.2em]

5. If the question is not answerable from the visible evidence, use mode = \dyjsonval{unsupported}.\\[0.4em]

Rules for evidence boxes:\\
- Each evidence item should localize a concrete visual region.\\
- Use normalized box coordinates.\\
- Avoid unrelated background regions or decorative objects.\\
- Multiple boxes and multiple frames are allowed when they support the answer.\\[0.4em]

Rules for rationale:\\
- Use a rationale only for mode = \dyjsonval{need\_reasoning}.\\
- Write one concise sentence describing the reasoning over grounded objects or events.\\
- Do not mention frame IDs, box IDs, coordinates, or annotation metadata.\\
- Do not reveal the answer by saying \dyjsonval{the answer is A/B/C/D} inside the rationale.\\[0.4em]

Output strict JSON:\\
\{\\
\hspace*{1em}\dyjsonkey{mode}: \dyjsonval{perception\_only} | \dyjsonval{need\_reasoning} | \dyjsonval{unsupported},\\
\hspace*{1em}\dyjsonkey{evidence}: [\\
\hspace*{2em}\{\\
\hspace*{3em}\dyjsonkey{frame\_id}: integer,\\
\hspace*{3em}\dyjsonkey{timestamp}: float,\\
\hspace*{3em}\dyjsonkey{entity}: \dyjsonval{short visual phrase},\\
\hspace*{3em}\dyjsonkey{bbox}: [x1, y1, x2, y2],\\
\hspace*{3em}\dyjsonkey{role}: \dyjsonval{evidence role}\\
\hspace*{2em}\}\\
\hspace*{1em}],\\
\hspace*{1em}\dyjsonkey{rationale}: \dyjsonval{empty for perception\_only},\\
\hspace*{1em}\dyjsonkey{answer}: \dyjsonval{answer label or text}\\
\}
\end{dypromptbox}

\subsection{Three-Stage Verification Prompts}

We verify generated annotations with three lightweight checks. 
The first two phases operate at the evidence-box level, while the third phase verifies rationale faithfulness for \textsc{reason} examples.

\paragraph{Phase A: Gold-blind crop usefulness.}
For each generated box, we crop the corresponding region with padding. 
The verifier sees the question and the cropped regions, but not the gold answer. 
This phase filters boxes that are empty, blurred, or irrelevant to the question.

\begin{dypromptbox}{Phase A Verification Prompt}
Question: \{question\}\\[0.4em]

You are shown cropped image regions, each labeled with a box ID.\\[0.4em]

For each crop, judge whether it contains useful visual evidence for answering the question.\\[0.4em]

Mark false only if:\\
1. the crop is empty, blurred, or contains no question-relevant content;\\
2. the crop is unrelated to the question.\\[0.4em]

Otherwise mark true.\\[0.4em]

Output strict JSON:\\
\{\\
\hspace*{1em}\dyjsonkey{judgments}: [\\
\hspace*{2em}\{\\
\hspace*{3em}\dyjsonkey{bbox\_id}: integer,\\
\hspace*{3em}\dyjsonkey{evidence\_useful}: true | false,\\
\hspace*{3em}\dyjsonkey{confidence}: float between 0 and 1,\\
\hspace*{3em}\dyjsonkey{reason}: \dyjsonval{at most 20 words}\\
\hspace*{2em}\}\\
\hspace*{1em}]\\
\}
\end{dypromptbox}

\paragraph{Phase B: Answer support verification.}
In Phase B, the verifier sees the question, the gold answer, and the original frames with the Phase-A surviving boxes overlaid. 
This phase checks whether each box supports the correct answer in its frame context.

\begin{dypromptbox}{Phase B Verification Prompt}
Question: \{question\}\\
Correct answer: \{gold\_answer\}\\[0.4em]

You are shown video frames with numbered bounding boxes.\\[0.4em]

For each numbered box, judge whether the content inside the box directly supports the correct answer.\\[0.4em]

Use the rest of the frame as spatial or contextual reference if the question involves position, comparison, or relationships.\\[0.4em]

If a temporal question requires multiple frames, give benefit of the doubt to a box that contributes necessary partial evidence.\\[0.4em]

Output strict JSON:\\
\{\\
\hspace*{1em}\dyjsonkey{judgments}: [\\
\hspace*{2em}\{\\
\hspace*{3em}\dyjsonkey{bbox\_id}: integer,\\
\hspace*{3em}\dyjsonkey{supports\_answer}: true | false,\\
\hspace*{3em}\dyjsonkey{confidence}: float between 0 and 1,\\
\hspace*{3em}\dyjsonkey{reason}: \dyjsonval{at most 20 words}\\
\hspace*{2em}\}\\
\hspace*{1em}]\\
\}
\end{dypromptbox}

\paragraph{Phase C: Rationale faithfulness.}
Phase C is applied only to \textsc{reason} examples. 
The verifier receives the question, gold answer, generated rationale, and key frames with evidence boxes overlaid. 
It checks whether the rationale is logical, visually grounded, and free from answer leakage or hallucinated visual content.

\begin{dypromptbox}{Phase C Verification Prompt}
Question: \{question\}\\
Correct answer: \{gold\_answer\}\\
Rationale: \{rationale\}\\[0.4em]

Below are key video frames with evidence boxes around relevant entities.\\[0.4em]

Judge whether the rationale faithfully reflects the visible evidence.\\[0.4em]

Check:\\
1. \dyjsonkey{logical\_to\_answer}: the reasoning chain validly leads to the correct answer.\\
2. \dyjsonkey{visually\_grounded}: the entities, actions, and events mentioned in the rationale are visible in the provided frames.\\
3. \dyjsonkey{leaks\_answer\_letter}: the rationale directly states the answer label, such as \dyjsonval{the answer is A}.\\
4. \dyjsonkey{hallucinates\_objects}: the rationale describes objects, actions, or events not visible in the frames.\\[0.4em]

Output strict JSON:\\
\{\\
\hspace*{1em}\dyjsonkey{logical\_to\_answer}: true | false,\\
\hspace*{1em}\dyjsonkey{visually\_grounded}: true | false,\\
\hspace*{1em}\dyjsonkey{leaks\_answer\_letter}: true | false,\\
\hspace*{1em}\dyjsonkey{hallucinates\_objects}: true | false,\\
\hspace*{1em}\dyjsonkey{reason}: \dyjsonval{at most 40 words}\\
\}
\end{dypromptbox}

The verified annotations determine which supervised signals are applied during training. 
Verified evidence boxes provide object-level targets for perception latents, while verified rationales are used only in the explicit-rationale view for rationale-to-latent distillation. 
The latent placeholder positions are not supervised as ordinary text tokens.

\subsection{Human Verification of Annotation Quality}\label{app:human_verif}
As the curated training annotations provide auxiliary supervision for distillation rather than a main contribution of this work, we assess their reliability with a human audit on a stratified sample of 100 training annotations, covering both \textsc{direct} and \textsc{reason} examples, rather than full manual verification of the dataset. We check (i) whether the grounded evidence boxes correctly localize the queried visual content, and (ii) whether the rationale is consistent with the video and supports the recorded answer. The evidence boxes are judged correct in 91\% of the sampled annotations, and the rationales are judged reliable in 94\%. 

\section{Full GRPO Objective}
\label{app:grpo}
For completeness, we give the clipped GRPO objective used in Section~\ref{sec:learning}, with the importance ratio $r_{i,t}(\theta)$ defined in Eq.~\eqref{eq:rl_ratio}:
\begin{equation}
\begin{aligned}
J_{\mathrm{RL}}(\theta)
={}&
\mathbb{E}
\Bigg[
\frac{1}{G}
\sum_{i=1}^{G}
\frac{1}{|\mathcal{T}_i|}
\sum_{t\in\mathcal{T}_i}
\Big[
\min\Big(
r_{i,t}(\theta)\hat{A}_i,
\\
&\quad
\mathrm{clip}
\big(
r_{i,t}(\theta),
1-\epsilon,
1+\epsilon
\big)
\hat{A}_i
\Big)
\\
&\quad
-
\beta D_{\mathrm{KL}}
\left(
\pi_\theta
\Vert
\pi_{\mathrm{ref}}
\right)
\Big]
\Bigg],
\end{aligned}
\label{eq:rl_obj}
\end{equation}
where $\mathcal{T}_i$ is the set of policy-scored text tokens in rollout $o_i$, $\pi_{\theta_{\mathrm{old}}}$ is the old policy, $\pi_{\mathrm{ref}}$ is the frozen SFT model, $\epsilon$ is the clipping coefficient, $\beta$ is the KL coefficient, and $\hat{A}_i$ is the group-normalized advantage computed from the rollout rewards in Eq.~\eqref{eq:reward}:
\begin{equation}
\hat{A}_i
=
\frac{R(o_i) - \mathrm{mean}\left(\{R(o_j)\}_{j=1}^{G}\right)}{\mathrm{std}\left(\{R(o_j)\}_{j=1}^{G}\right)}.
\label{eq:advantage}
\end{equation}

\section{Training Details}
\label{app:hyper}
During SFT, we train for one epoch with AdamW at a learning rate of $1\times10^{-5}$. 
The number of perception latents follows the number of verified evidence boxes in each SFT example, while the number of reasoning latents is fixed at $K_r=6$. 
The loss weights in Eq.~\eqref{eq:sft_loss} are $\lambda_g=1.0$, $\lambda_e=1.0$, and $\lambda_d=20.0$, with $\lambda_d$ following CODI's distillation weight~\citep{shen2025codi}. 
During RL, we use GRPO with learning rate $5\times10^{-7}$, group size $G=8$, sampling temperature $\tau=0.9$, clipping coefficient $\epsilon=0.2$, and KL coefficient $\beta=0.04$ for 2{,}500 steps. 
Rollouts use the inference-time latent budgets $K_p=4$ and $K_r=6$, and the reward coefficients in Eq.~\eqref{eq:reward} are $\alpha_{\mathrm{form}}=0.05$ and $\alpha_{\mathrm{acc}}=1.0$.

\section{Benchmark Details}
\label{app:benchmarks}
All locally evaluated baselines and \methodname{} share one visual protocol: 16 uniformly sampled frames per video, each capped at $307{,}200$ pixels, using the prompt template and decoding configuration released in each model's official repository. 

CoT-prompted and Thinking base models sometimes fail to follow the instructed answer format; for these outputs, we use an LLM judge to identify the selected option.
Because we use a unified frame budget and the full official question sets, numbers in Table~\ref{tab:main} are not directly comparable to those originally reported. 
We evaluate on nine benchmarks in total; all questions are presented as lettered multiple-choice QA, and unless otherwise noted we evaluate the full official set.

\textbf{Video-MME}~\citep{fu2025video} covers general video understanding with 2{,}700 four-option questions over 900 videos, evenly split into short, medium, and long duration tiers (900 questions each). 
\textbf{LVBench}~\citep{wang2025lvbench} targets hour-scale long-video comprehension with 1{,}549 four-option questions over 103 videos. 
\textbf{LongVideoBench}~\citep{wu2024longvideobench} poses referring-reasoning questions over long videos; we evaluate the full validation split of 1{,}337 questions, as the test split withholds ground truth. 
\textbf{MVBench}~\citep{li2024mvbench} spans 20 temporal perception task types with 200 questions each (4{,}000 in total). 
\textbf{LongVideo-Reason}~\citep{chen2025scalingrllongvideos} focuses on reasoning over long videos, with a curated evaluation set of 1{,}000 questions. 
\textbf{TempCompass}~\citep{liu2024tempcompassvideollmsreally} probes temporal understanding with 7{,}540 instances spanning its four official task formats; we follow Video-R1's official evaluation set~\citep{NEURIPS2025_8eb39768}, in which all instances are cast as lettered multiple-choice questions. 
\textbf{Video-TT}~\citep{zhang2025videothinkingtestholistic} is a holistic video thinking test; we evaluate its multiple-choice split of 1{,}000 five-option questions (one per video). 
\textbf{Video-Holmes}~\citep{cheng2025video} requires multi-step inference over suspense short films, with 1{,}837 six-option questions. 
\textbf{MMVU}~\citep{zhao2025mmvumeasuringexpertlevelmultidiscipline} evaluates expert-level, multi-discipline video understanding; we evaluate the 625 multiple-choice questions of its public validation split, with the remaining 375 open-ended questions excluded.

\section{More Ablation Result}\label{RL_overall}
\subsection{RL Ablation Across Four Backbones}
Table~\ref{tab:rl_all} reports the per-backbone breakdown of the SFT$\to$RL comparison summarised in Section~\ref{sec:ablation}, where RL improves the nine-benchmark average on all four backbones by $+1.1$ to $+1.5$ points.
\begin{table*}[t]
\centering
\caption{
Per-backbone view of the RL stage.
\textbf{Bold} marks the cells RL improves; $\Delta$ is the change in the nine-benchmark average.
}
\label{tab:rl_all}
\resizebox{\textwidth}{!}{
\begin{tabular}{l ccccc cccc c r}
\toprule
\textbf{Backbone} & Video-MME & LVBench & MVBench & MMVU & LongVB & LVR & TempC & VHolmes & VTT & \textbf{Avg.} & \textbf{$\Delta$} \\
\midrule
\textbf{Qwen2.5-VL-7B} & 58.7$\to$\textbf{59.8} & 40.6$\to$\textbf{41.2} & 65.6$\to$\textbf{66.3} & 61.4$\to$\textbf{64.5} & 55.8$\to$\textbf{57.4} & 73.2$\to$\textbf{74.6} & 71.8$\to$\textbf{72.4} & 39.8$\to$\textbf{41.0} & 37.4$\to$\textbf{40.0} & 56.0$\to$\textbf{57.5} & \textbf{+1.5} \\
\textbf{Qwen3-VL-4B} & 58.2$\to$\textbf{59.6} & 40.5$\to$39.8 & 65.2$\to$\textbf{66.7} & 65.4$\to$63.8 & 57.7$\to$\textbf{58.9} & 76.4$\to$\textbf{76.9} & 66.5$\to$\textbf{69.6} & 43.2$\to$\textbf{47.5} & 38.3$\to$\textbf{40.7} & 56.8$\to$\textbf{58.2} & \textbf{+1.4} \\
\textbf{InternVL3.5-4B} & 56.4$\to$\textbf{57.6} & 37.7$\to$\textbf{38.5} & 66.1$\to$\textbf{67.2} & 62.1$\to$\textbf{62.7} & 55.9$\to$\textbf{56.8} & 74.5$\to$73.9 & 68.5$\to$\textbf{70.7} & 40.8$\to$\textbf{44.7} & 38.6$\to$38.2 & 55.6$\to$\textbf{56.7} & \textbf{+1.1} \\
\textbf{LLaVA-OneVision-7B} & 55.0$\to$\textbf{55.7} & 36.8$\to$\textbf{37.3} & 60.4$\to$\textbf{61.0} & 54.4$\to$\textbf{56.6} & 54.6$\to$\textbf{55.8} & 72.1$\to$\textbf{75.1} & 65.1$\to$\textbf{66.0} & 43.0$\to$42.0 & 35.4$\to$\textbf{38.0} & 53.0$\to$\textbf{54.2} & \textbf{+1.2} \\
\bottomrule
\end{tabular}
}
\end{table*}

\subsection{SFT Objective Ablation under Different Frames}
\label{app:sft_frames}

Tables~\ref{tab:ablation_sft_lvb_f32} and~\ref{tab:ablation_sft_lvb_f64} repeat the SFT objective ablation of Table~\ref{tab:ablation_sft_lvb} at 32- and 64-frame test-time budgets, where the full \methodname{} objective again attains the highest overall accuracy.

\begin{table*}[t]
\centering
\small
\caption{
SFT objective ablation on LVBench with Qwen3-VL-4B at a 32-frame budget.
``Perception latent + text CoT'' keeps the perception latent block but verbalises the rationale;
``Text-CoT SFT'' performs all reasoning in the vocabulary space;
``Answer-only SFT'' removes both latent blocks and supervises only the final answer.}
\label{tab:ablation_sft_lvb_f32}
\resizebox{0.9\textwidth}{!}{
\begin{tabular}{l cccccc c}
\toprule
\textbf{Variant} & Entity & Event & Key Info. & Reason & Temporal & Summ. & \textbf{Overall} \\
\midrule
\multicolumn{8}{l}{\emph{Text-reasoning SFT baselines}} \\
Perception latent + text CoT
& 42.5 & \second{41.4} & 42.3 & 40.3 & 34.5 & \second{34.5} & 41.6 \\
Text-CoT SFT
& \second{44.0} & 40.3 & 40.2 & \second{41.3} & 33.2 & 31.0 & \second{41.8} \\
Answer-only SFT
& 40.6 & 40.5 & 40.5 & 39.3 & 34.1 & 32.8 & 40.5 \\
\midrule
\multicolumn{8}{l}{\emph{\methodname{} latent SFT}} \\
\textbf{\methodname{} SFT (full)}
& \textbf{44.2} & \textbf{42.2} & 41.6 & 38.8 & 34.5 & \second{34.5} & \textbf{42.1} \\
\quad w/o reasoning latents
& 42.5 & 40.5 & 40.5 & 40.3 & \second{37.7} & \textbf{37.9} & 41.1 \\
\quad w/o reasoning supervision
& 42.5 & 40.3 & 42.3 & 38.3 & \textbf{38.6} & 32.8 & 40.9 \\
\quad w/o perception grounding
& 42.5 & 39.7 & \second{42.6} & \textbf{44.3} & 36.4 & 31.0 & 41.4 \\
\quad w/o adaptive routing (mandatory latent reasoning)
& 42.2 & 40.5 & \textbf{43.6} & 40.8 & 33.2 & \second{34.5} & 41.3 \\
\bottomrule
\end{tabular}
}
\end{table*}

\begin{table*}[t]
\centering
\small
\caption{
SFT objective ablation on LVBench with Qwen3-VL-4B at a 64-frame budget, before RL. ``Perception latent + text CoT'' keeps the perception latent block but verbalises the rationale;
``Text-CoT SFT'' performs all reasoning in the vocabulary space;
``Answer-only SFT'' removes both latent blocks and supervises only the final answer.}
\label{tab:ablation_sft_lvb_f64}
\resizebox{0.9\textwidth}{!}{
\begin{tabular}{l cccccc c}
\toprule
\textbf{Variant} & Entity & Event & Key Info. & Reason & Temporal & Summ. & \textbf{Overall} \\
\midrule
\multicolumn{8}{l}{\emph{Text-reasoning SFT baselines}} \\
Perception latent + text CoT
& 44.6 & 41.3 & 43.6 & 42.3 & 34.1 & \second{34.5} & 42.8 \\
Text-CoT SFT
& \second{47.3} & 42.0 & 44.0 & 40.3 & 34.1 & 29.3 & 43.8 \\
Answer-only SFT
& 46.4 & 41.4 & 47.8 & \second{43.8} & 36.8 & \textbf{41.4} & 44.0 \\
\midrule
\multicolumn{8}{l}{\emph{\methodname{} latent SFT}} \\
\textbf{\methodname{} SFT (full)}
& \textbf{48.6} & \textbf{44.2} & \second{48.5} & 42.8 & 33.6 & 31.0 & \textbf{45.3} \\
\quad w/o reasoning latents
& 46.5 & \second{42.8} & 45.0 & \textbf{45.3} & \second{38.2} & 27.6 & \second{44.1} \\
\quad w/o reasoning supervision
& 44.2 & 41.9 & 47.8 & 41.3 & \textbf{39.1} & 32.8 & 43.2 \\
\quad w/o perception grounding
& 47.1 & 41.4 & \textbf{49.5} & 41.8 & 37.3 & 29.3 & 43.8 \\
\quad w/o adaptive routing (mandatory latent reasoning)
& 47.0 & 41.7 & 46.0 & 41.3 & 33.6 & 24.1 & 43.0 \\
\bottomrule
\end{tabular}
}
\end{table*}

\subsection{Inference Frame Budget}
\label{app:frame_budget}

Table~\ref{tab:frame_budget} reports the per-benchmark breakdown behind the frame-budget analysis in Section~\ref{sec:ablation}, where \methodname{} stays ahead of Qwen3-VL-4B-Thinking at 16, 32, and 64 frames.

\begin{table*}[t]
\centering
\caption{
Effect of the inference frame budget on the nine main benchmarks. LongVB: LongVideoBench. LVR: LongVideo-Reason. TempC: TempCompass. VHolmes: Video-Holmes. VTT: Video-TT.
}
\label{tab:frame_budget}
\resizebox{\textwidth}{!}{
\begin{tabular}{l ccccc cccc c}
\toprule
\textbf{Model} & Video-MME & LVBench & MVBench & MMVU & LongVB & LVR & TempC & VHolmes & VTT & \textbf{Avg.} \\
\midrule
\multicolumn{11}{l}{\emph{16 frames}} \\
Qwen3-VL-4B-Thinking
& 56.3 & 36.9 & 61.0 & 66.7 & 56.3 & 66.8 & 71.3 & 36.7 & 34.0 & 54.0 \\
\methodname{} (Qwen3-VL-4B)
& 59.6 & 39.8 & 66.7 & 63.8 & 58.9 & 76.9 & 69.6 & 47.5 & 40.7 & 58.2 \\
\midrule
\multicolumn{11}{l}{\emph{32 frames}} \\
Qwen3-VL-4B-Thinking
& 60.9 & 37.5 & 62.4 & 68.3 & 57.9 & 71.1 & 72.0 & 40.3 & 36.0 & 56.3 \\
\methodname{} (Qwen3-VL-4B)
& 62.3 & 43.4 & 66.7 & 63.0 & 62.0 & 77.6 & 70.0 & 49.4 & 42.5 & 59.7 \\
\midrule
\multicolumn{11}{l}{\emph{64 frames}} \\
Qwen3-VL-4B-Thinking
& 63.3 & 40.7 & 61.6 & 68.6 & 61.8 & 72.4 & 71.1 & 42.5 & 37.6 & 57.7 \\
\methodname{} (Qwen3-VL-4B)
& 63.5 & 45.7 & 65.9 & 64.0 & 63.2 & 79.5 & 69.1 & 50.4 & 43.3 & 60.5 \\
\bottomrule
\end{tabular}
}
\end{table*}

\subsection{Matched Low-Resolution Protocol}
\label{app:lowres}

Table~\ref{tab:lowres} reports the per-benchmark breakdown behind the pixel-budget analysis in Section~\ref{sec:ablation}, where \methodname{} attains the highest nine-benchmark average of the three models under the matched protocol.

\begin{table*}[t]
\centering
\caption{
Per-benchmark results under the matched low-resolution protocol
($128\times28\times28$ for training, $256\times28\times28$ for evaluation). LongVB: LongVideoBench. LVR: LongVideo-Reason. TempC: TempCompass. VHolmes: Video-Holmes. VTT: Video-TT.
}
\label{tab:lowres}
\resizebox{\textwidth}{!}{
\begin{tabular}{l ccccc cccc c}
\toprule
\textbf{Model} & Video-MME & LVBench & MVBench & MMVU & LongVB & LVR & TempC & VHolmes & VTT & \textbf{Avg.} \\
\midrule
\textbf{Video-R1} & 57.7 & 34.3 & 63.6 & 65.4 & 55.1 & 71.3 & 70.0 & 40.8 & 41.6 & 55.5 \\
\textbf{VideoRFT} & 55.7 & 33.0 & 62.0 & 66.2 & 53.3 & 70.5 & 70.7 & 40.8 & 41.0 & 54.8 \\
\textbf{\methodname{} (Qwen2.5-VL-7B)} & 58.9 & 39.7 & 67.9 & 62.4 & 56.8 & 74.1 & 73.1 & 42.2 & 36.6 & \textbf{56.9} \\
\bottomrule
\end{tabular}
}
\end{table*}

\section{More Visualizations}
\label{app:vis}
\begin{figure*}[t]
\centering
\includegraphics[width=0.9\textwidth]{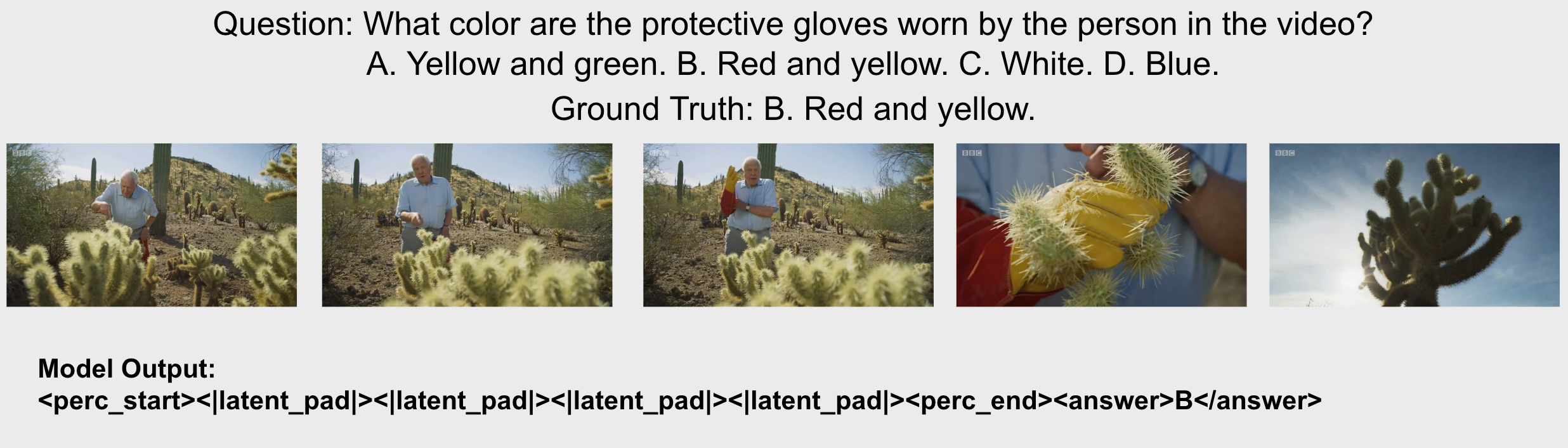}
\caption{
\methodname{} distinguished this case does not require reasoning, as the color of the glove is obviously shown in the image. 
}
\label{fig:case_per}
\end{figure*}

\begin{figure*}[t]
\centering
\includegraphics[width=0.9\textwidth]{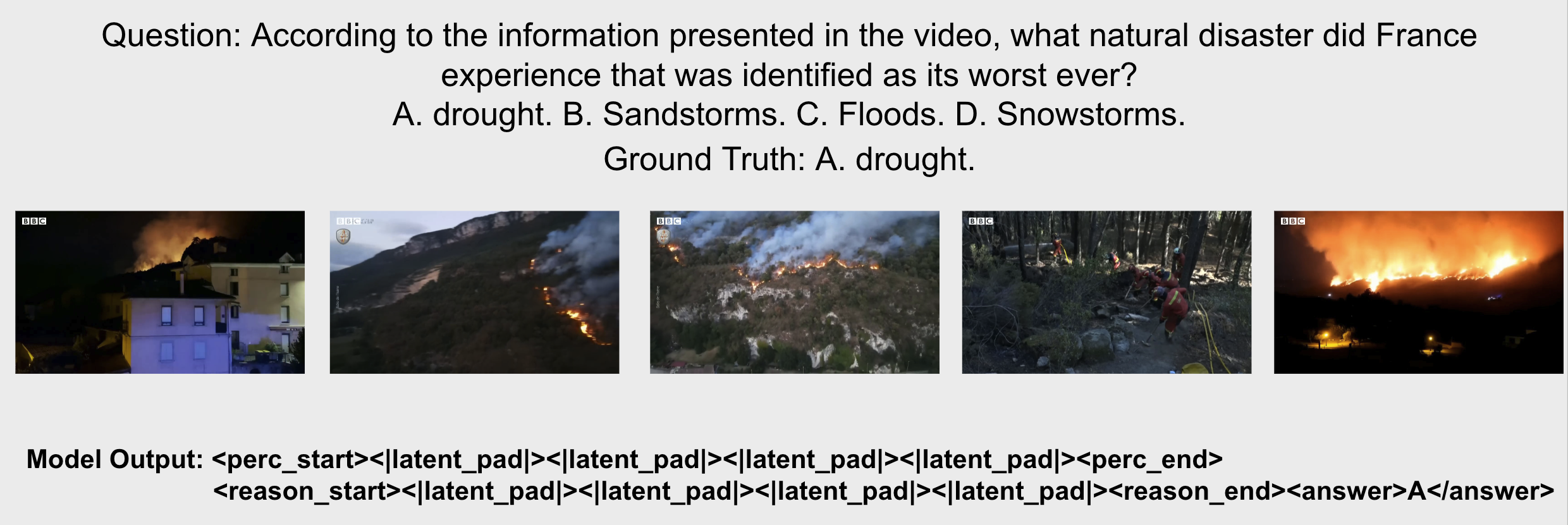}
\caption{
\methodname{} routes this case through its reasoning segment: the frames show its downstream wildfires—rather than the disaster itself. The model infers the visible evidence into the underlying cause (drought). 
}
\label{fig:case_rea}
\end{figure*}

In this section, Figure~\ref{fig:case_per} and  Figure~\ref{fig:case_rea} show two routes of our DyLaR model. The model distinguished the task successfully and therefore entered the correct latent route.


\end{document}